\documentclass{article}
\usepackage[dblblindworkshop,final]{neurips_2026}
\workshoptitle{Linguistic Principles for Foundation Models (LP4FM)}
\usepackage[utf8]{inputenc}
\usepackage[T1]{fontenc}
\usepackage{hyperref}
\usepackage{url}
\usepackage{booktabs}
\usepackage{amsfonts}
\usepackage{amsmath}
\usepackage{nicefrac}
\usepackage{microtype}
\usepackage{graphicx}
\hypersetup{pdftitle={The Score Is Not the Structure: Brain Alignment and Cross-Lingual Transfer},pdfauthor={Saman Rahbar}}

\title{The Score Is Not the Structure:\\
Brain Alignment and Cross-Lingual Transfer}

\author{%
  Saman Rahbar\\
  Independent researcher\\
  Vancouver, Canada\\
  \texttt{info@srahbar.com}\\
  \url{https://srahbar.com}
}

\begin{document}
\maketitle

\begin{abstract}
Similarity scores are often offered as evidence that a model shares structure with the brain or across languages. We ask what such a score reads when that structure is removed, or when the instrument measuring it does not work, in two settings. Across seventeen languages, a grammaticality probe transfers worse between more distant languages ($r=-0.66$). But the probe's own accuracy falls along the same axis and is at chance in four languages, four of the five most distant. Dropping them halves the explained variance, and because it also narrows the range of distances, the design cannot say how much of the gradient is the instrument. Counting the $272$ language pairs as independent gives $p=0.0006$ for a steering effect that is null when the seventeen languages are the unit ($p=0.155$). In brain alignment, a training objective raises a language model's similarity to fMRI responses from $0.10$ to $0.34$ (reliability ceiling $0.54$), yet targets with the brain correspondence destroyed still reach $0.31$. A shuffled target scores about $k/n$ against the real one, for rank $k$ and $n$ sentences, a baseline that can be computed before any model is trained. Both interventions work, yet we detect no distance-graded steering effect and no syntactic benefit from alignment. Before reading a correspondence score, measure how much of it survives without the correspondence.
\end{abstract}

\section{Introduction}
Representational similarity has become a common currency for relating models to one another and to the brain. Language models are scored by how well their activations predict neural recordings; cross-lingual models are probed for shared subspaces; interpretability work reads structure off learned directions. In each case it is tempting to infer that measured \emph{correspondence}, to the brain, across languages, or to some reference geometry, is evidence of a functionally meaningful, and therefore usable, property. The inference shows up as a research program: if the correspondence is real, perhaps it can be turned into a training signal that improves the model, or into a causal handle that steers its behavior.

That inference has a step before it that is rarely taken. A correspondence score is a measurement, and like any measurement it can register a quantity that is not the one intended. Three failure modes matter here. A score can be \emph{insensitive}, returning nearly the same value whether or not the reference structure is present, in which case its magnitude carries little about correspondence. The \emph{instrument} used to obtain it can vary in reliability across the very axis being studied, in which case the resulting gradient partly tracks measurement quality instead of the phenomenon. And the \emph{units} can be miscounted, when a design generates many observations from few independent entities. None is visible from the score alone, and each admits a direct test.

We apply those tests in two settings, then ask the utility question of whatever survives them. The settings are deliberately different, one comparing a shared direction across languages and one aligning a model to the brain, because a methodological point demonstrated in two independent regimes is worth more than either case alone. Which failure dominates differs between them, which is part of the argument: the typological gradient is compromised mainly by its instrument, the alignment score mainly by insensitivity.

In neither setting does the score mean what its size suggests. Our contributions are:
\begin{itemize}
\item an audit of three inexpensive checks, run before any question about usefulness: remove the correspondence and re-score, check that the instrument is equally reliable everywhere it is compared, and count the independent units correctly (Section~\ref{sec:framework});
\item a cross-lingual case in which probe accuracy falls with typological distance ($r=-0.74$) and is at chance in four of seventeen languages, so the transfer gradient cannot be separated from measurement quality, and in which counting pairs rather than languages would have turned a null into $p=0.0006$ (Section~\ref{sec:xling});
\item a brain-alignment case in which targets with the correspondence destroyed recover $83$--$92\%$ of the alignment score, and in which the target-to-target baseline follows $k/n$, a quantity anyone can compute in advance (Section~\ref{sec:brain});
\item interventions shown to work before their usefulness is tested, after which neither correspondence yields a detectable benefit in these settings.
\end{itemize}
Code, data processing and raw results for both studies are at \url{https://github.com/saman-rahbar/score-is-not-the-structure}.

\section{Related work}\label{sec:related}
Model--brain alignment is most often reported as a post-hoc evaluation of correspondence and not as a training objective \citep{Zhang2026}, and brain-inspired redesigns \citep{Zhao2025,Pezzulo2026} and inductive-bias studies \citep{Papadimitriou2023,Feinman2018,Dijujin2025} argue that cognitively motivated structure helps without testing a brain-derived signal under matched controls. Cross-lingually, multilingual models encode shared subject-verb agreement structure \citep{mueller2020}, and downstream transfer tracks typological distance captured by URIEL/lang2vec features \citep{littell2017,lin2019}, although in lexical normalization the best number of co-trained languages shows no reliable relation to typological distance \citep{Rahbar2026curse}; activation steering manipulates behavior along learned directions \citep{turner2023}.

Our audit borrows its logic from three places. \citet{hewitt2019} show that a probe's accuracy is uninterpretable without a control task establishing what the same probe achieves on structure it should not find, and shuffled-label controls have been used in the same way to check that layer-wise probe transitions in language models track linguistic structure \citep{Rahbar2026ignition}; selectivity is a within-language property, asking whether a probe reads the representation or memorizes the labels. Our instrument check is orthogonal to it. A probe can be perfectly selective in every language and still differ in absolute reliability between them, and it is that between-language variation, not memorization, that biases a gradient fitted across languages. Selectivity licenses the claim that a probe measures something; comparability licenses the claim that two probes measure it equally well, which is what a cross-lingual regression assumes. We did not find the check reported in the work we build on \citep{littell2017,lin2019,mueller2020}, though we have not surveyed the literature systematically enough to claim it is never done.

\citet{kornblith2019} characterize what CKA is and is not sensitive to as a similarity measure; we ask the complementary empirical question of how much of a fitted CKA score survives destroying the correspondence it scores. A similar negative control in brain-imaging foundation models, probing an untrained copy of the model and the raw input, shows that scanner site is decodable before any pretraining \citep{Rahbar2026site}. The unit-counting problem is older still: \citet{mantel1967} introduced the matrix permutation that comparative biology adopted for exactly the non-independence that pairwise language comparisons inherit.

\section{Auditing a correspondence measure}\label{sec:framework}
Both studies follow one template. Let a model carry an internal structure $S$ (a layer's representational geometry; a linear direction) that we bring into correspondence with a reference $R$ (fMRI response geometry; another language's grammaticality contrast), and let $C(S,R)$ be the score reporting that correspondence.

The audit has three parts, taken in this order because each makes the next interpretable. (i) A \emph{content ablation} rebuilds the reference so that it keeps every incidental property but loses the correspondence itself. We use three: a target whose rows are shuffled, which preserves its statistics and shape but lines it up with nothing; noise of the same dimension, which keeps the training pressure identical; and a random direction the same size as the real one. Re-measuring $C$ against these gives the share of the score attributable to content rather than to form, rank, or scale. (ii) An \emph{instrument check} asks whether $C$ is estimated with comparable reliability everywhere it is compared. Where $C$ is read through a fitted probe, a probe that fails within a language cannot support a claim about that language's transfer; where reliability co-varies with the predictor, as it does with typological distance, the resulting gradient mixes signal with measurement quality. (iii) An \emph{inference check} counts the units correctly. Pairwise designs generate many observations from few entities, so significance must be assessed by permuting the entities and not the pairs.

Both studies carry out all three, under names their own literatures use. The cross-lingual study's content ablation is the random-direction steering baseline, matched in perturbation magnitude and carrying no source-specific content; its instrument check is within-language probe accuracy; its inference check is the language-level permutation. The brain-alignment study's content ablation is the shuffled and random target. Its instrument check is the subject-split noise ceiling, which asks whether the target supports a reliable CKA at all before any is attributed to the model; its inference check is pairing over shared seeds. What differs between the studies is which part proves decisive, and that is itself informative: the instrument dominates in the cross-lingual study, the ablation in the brain-alignment study.

Only after these do we ask the functional question: does exploiting the surviving correspondence change a downstream behavior $B$ (syntactic generalization; grammatical preference) beyond matched controls? Running the audit first is what keeps a null on $B$ interpretable, since a null is uninformative if the intervention never installed the content, and a positive result is uninformative if the score never measured it. Table~\ref{tab:audit} summarizes each check and what it found in both studies.

\begin{table}[t]
\centering
\caption{The audit in both studies: what each check is, and what it found.}
\label{tab:audit}
\small
\setlength{\tabcolsep}{4pt}
\begin{tabular}{p{0.15\linewidth}p{0.39\linewidth}p{0.39\linewidth}}
\toprule
Check & Cross-lingual transfer (Section~\ref{sec:xling}) & Brain alignment (Section~\ref{sec:brain}) \\
\midrule
Content ablation & Random steering direction of equal size: real directions beat it for all seventeen source languages, and once it is subtracted no distance trend remains & Shuffled and noise targets: they recover $83$--$92\%$ of the alignment score \\
Instrument check & Within-language probe accuracy: at chance in four of seventeen languages, and falling with distance ($r=-0.74$) & Split-half noise ceiling: the target is reliable ($0.54$) \\
Inference check & Permute languages, not pairs: steering trend $p=0.0006 \to 0.155$ & Pair conditions over shared seeds: brain-specific increment $+0.03$ to $+0.07$, $p<0.01$ \\
Usefulness & No distance-graded steering effect detected & No BLiMP gain detected \\
\bottomrule
\end{tabular}
\end{table}
\section{Study 1: Cross-lingual transfer and steering}\label{sec:xling}
\textbf{Setup.} In a multilingual autoregressive model, XGLM-1.7B \citep{lin2021xglm}, chosen because it covers all seventeen languages, we fit a \emph{probe}: a logistic-regression classifier ($L_2$, $C{=}1$, on standardized features) trained to tell grammatical from ungrammatical subject-verb agreement. It reads the residual stream at the middle decoder layer, at the position of the verb that carries the agreement, and is trained on $60\%$ of each language's pairs and tested on the rest. The training data are \emph{minimal pairs}, sentences identical except for the agreement error, taken from MultiBLiMP \citep{jumelet2025} for seventeen languages that mark agreement on the verb.

For every ordered pair of languages $(A,B)$ we measure two things. \emph{Transfer} is how accurate $A$'s probe is when applied to $B$. \emph{Steering} is what happens when we add $A$'s direction into the model while it reads $B$, and ask whether $B$'s preference for the grammatical sentence strengthens. We measure that preference as the \emph{margin}: the model's total log-probability of the grammatical sentence minus that of its ungrammatical twin, in nats. A margin of zero means the model is indifferent between them, so the steering effect of $+6.21$ reported below is a shift of about six nats. We add the direction at the same layer the probe reads, at two strengths scaled to the size of the activations, and average the two. The random directions we compare against are scaled the same way, so they differ from the real one only in where they point. We regress both on the URIEL/lang2vec syntactic distance between $A$ and $B$ ($272$ pairs). The causal claim under test is that the \emph{steering} benefit decays with typological distance in a way not reducible to the correlational probe.

\begin{figure}[t]
\centering
\includegraphics[width=0.94\linewidth]{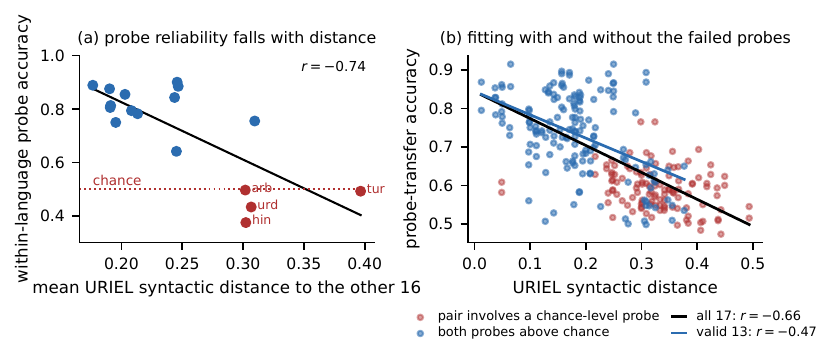}
\caption{The instrument varies along the axis under study. \emph{(a)} Within-language probe accuracy against each language's mean URIEL syntactic distance to the other sixteen, one point per language; red marks the four at or below chance, where a transfer accuracy has nothing to be transferred from. Reliability and the predictor co-vary at $r=-0.74$. \emph{(b)} Probe transfer against distance over all $272$ ordered pairs. Fitting all seventeen languages gives $r=-0.66$; fitting only the thirteen whose probe clears chance gives $r=-0.47$, halving the explained variance. Because the excluded four are also among the most distant, that difference cannot be attributed to validity alone; see Section~\ref{sec:xling}.}
\label{fig:validity}
\end{figure}

\textbf{Transfer falls with typological distance.} Probe transfer behaves as a shared-representation account predicts: transfer accuracy decreases with typological distance, $r=-0.66$ ($R^2{=}0.43$), reproducing the relationship that motivates URIEL-based transfer prediction.

\textbf{So does the probe's own accuracy.} A transfer accuracy is only interpretable where the probe it is transferred from works. In four of the seventeen languages the probe does no better than guessing on its own language: Hindi $0.375$, Urdu $0.433$, Turkish $0.493$, Arabic $0.497$, against a chance level of $0.5$. These failures are not scattered at random. A language's probe gets worse the more it differs from the rest of the sample, at $r=-0.74$ (Figure~\ref{fig:validity}a). We measure each language's distance as its mean distance to the other sixteen, so that no single language acts as the reference point. Probe reliability and the predictor of interest therefore co-vary, which is the instrument failure of Section~\ref{sec:framework} in its most consequential form. Restricting to the thirteen languages whose probe clears chance at both endpoints ($156$ pairs) halves the explained variance: $r=-0.47$, $R^2{=}0.22$ (Figure~\ref{fig:validity}b).

Bootstrapping over languages, the unit that matters here, rather than over pairs, the full-cohort correlation stays well away from zero at $[-0.79,-0.50]$ and the reliability gradient at $[-0.89,-0.51]$, but the restricted one spans zero at $[-0.77,+0.18]$. A permutation test still rejects no-association for it ($p{=}0.025$), since permutation and interval answer different questions, but thirteen units support the direction far better than the magnitude.

These are not simply undertrained probes. We cap each language at $300$ minimal pairs and train on $60\%$, so sixteen of the seventeen fit on exactly $180$ examples. Differences in training size therefore cannot explain why four of them fail, since there are almost none to speak of. (Correlating accuracy against the amount used gives $r=-0.04$, but with the variance removed by design that number carries little on its own.) What the cap cannot rule out is an effect of the absolute level: perhaps $180$ examples suffice for some languages and not others. One case speaks against that too, though only one. Basque trains on the fewest examples of any language, $163$, and still reaches $0.755$, above all four failures. Whatever makes Hindi, Urdu, Turkish and Arabic fail, it is not that they had less data.

\textbf{Dropping the failed probes also narrows the range of distances.} The four failing languages rank first, third, fourth and fifth of seventeen by mean typological distance, which is what $r=-0.74$ implies. Removing them therefore truncates the predictor's range, and truncation attenuates a correlation whatever the probes were doing. The two explanations are not separable in this sample. Of all $2{,}380$ four-language exclusions, only three others restrict the range as much as the invalid set does: two attenuate the correlation more ($r=-0.46$ and $-0.33$) and the third about as much ($-0.48$), against $-0.47$ for the invalid set. Dropping the four most distant \emph{valid} languages instead does not settle it either, since their summed distance ($1.047$) falls well short of the invalid set's ($1.307$): there is no distance-matched comparison to make. We therefore do not claim that half the gradient is measurement artifact. What we claim is weaker and, we think, more useful. On a typologically diverse sample, probe reliability and typological distance co-vary so strongly that the transfer gradient is \emph{not identifiable} without a control that separates them, and we are not aware of a study that reports one. That is a property of the design every such study shares, not of ours.

\begin{figure}[t]
\centering
\includegraphics[width=0.94\linewidth]{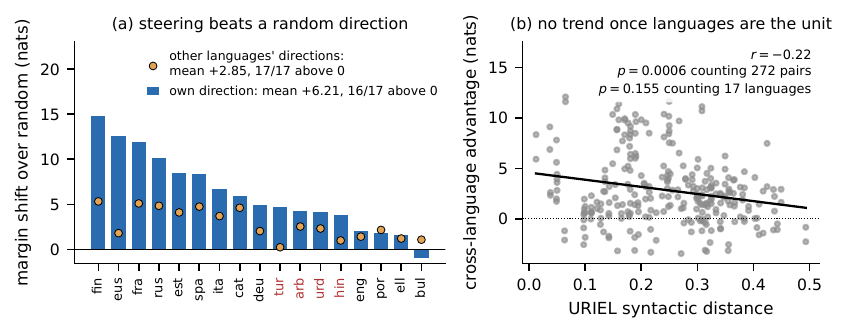}
\caption{Steering works, but not in proportion to distance. \emph{(a)} How much more a steering direction raises the grammatical-preference margin than a random direction of the same size: along each language's own direction (bars), and averaged over the sixteen targets when that language's direction steers the others (dots). Red marks the four languages whose probe is at chance. \emph{(b)} The cross-language advantage for all $272$ ordered pairs against URIEL distance. The fitted slope is significant if the pairs are counted as independent ($p=0.0006$), but not when the seventeen languages are the unit ($p=0.155$).}
\label{fig:steering}
\end{figure}

\textbf{Counting languages, not pairs.} The $272$ pairs come from seventeen languages, each appearing in $32$ of them, with phylogenetically structured distances, so a permutation over pairs treats as independent what is not. Permuting \emph{language labels} instead, the standard Mantel construction, the transfer gradient survives comfortably ($p{=}10^{-4}$; no permutation of $10{,}000$ reached the observed $r$). The steering result does not, as we show next.

\textbf{Steering works, within and across languages.} A null on how steering varies with distance means little unless steering does anything at all. The design as originally run could not say: its $272$ ordered pairs exclude self-pairs, so no language was ever steered along its own direction. Adding that condition, each language steered by its own full-data probe and contrasted against the per-target random baseline, the intervention moves the grammatical-preference margin by $+6.21$ on average, positive in sixteen of the seventeen languages (sign-flip $p \le 5\times10^{-5}$; Figure~\ref{fig:steering}a). It is also effective across languages, which is what the null below actually concerns. Pooled over all $272$ ordered pairs, steering $B$ by $A$'s direction beats the per-target random baseline by $+2.85$ nats, on $223$ of the $272$ pairs. Averaging within each source language gives a positive advantage for all seventeen (sign-flip $p{=}10^{-4}$). Cross-lingual steering is weaker than a language's own direction, as expected, but it plainly does something.

\textbf{Steering shows no trend with distance.} Raw steering benefit shows no relationship with distance ($r=+0.08$, $p{=}0.586$). A \emph{random}-direction steering baseline points the same way ($r=+0.24$), suggesting typologically unusual languages are simply more perturbable by any intervention, though under the language-level null this is itself only a trend ($p{=}0.081$). Removing that nuisance with a per-target random baseline, averaged over twenty directions, leaves a source-specific advantage that appears to decay with distance ($r=-0.22$) but does not survive the language-level null ($p{=}0.155$; Figure~\ref{fig:steering}b). Partialling out probe-transfer accuracy leaves nothing either, and with the sign reversed ($r=+0.11$, $p{=}0.455$). Restricted to the thirteen languages whose probe clears chance, the corrected relationship is $r=-0.09$ ($p{=}0.704$). No stage of this analysis produces a distance-graded steering effect once the seventeen languages, and not the $272$ pairs, are treated as the units.

We read this cautiously in one direction only. The effects are small, and partialling out a probe estimated with noise can perturb a weak partial correlation, so we do not claim that steering \emph{provably} carries no causal information; we make the weaker point that we find no evidence it is distance-graded. The unit correction changes what there is to explain: without it there would be an apparent effect to explain away, and the explaining away would itself have been the finding. With it, there is nothing to explain, and the within-language check above rules out the alternative reading that there was nothing to find. The shared direction is clearly real as a \emph{descriptor} (it transfers, $r=-0.66$, and $r=-0.47$ on the languages whose probe works), while its value as a distance-dependent causal \emph{lever} finds no support at any point.

\section{Study 2: Brain-geometry alignment and syntactic generalization}\label{sec:brain}
\textbf{Setup.} We fine-tune Pythia models \citep{biderman2023} (160M, 410M, 1.4B) on the stimulus text of the fMRI benchmark of \citet{pereira2018} ($627$ sentences), adding a second training objective that pulls the model's internal geometry toward the fMRI responses. The similarity measure is linear centered kernel alignment (CKA), which compares two sets of representations by how similarly they arrange the same sentences relative to one another, and returns $1$ for identical arrangements. Concretely, each training batch holds $16$ sentences. $H$ has one row per sentence, the model's layer-$6$ hidden states averaged over that sentence's tokens, and $B$ has the matching rows of the fMRI target. CKA compares how the batch's sentences are arranged relative to one another in the two spaces. The loss is $\mathcal{L}_{\mathrm{align}} = 1-\operatorname{CKA}(H,B)$ with weight $\lambda$, added to the usual next-token loss. We use Pythia because it offers three sizes trained on identical data. The fMRI target pools language-network voxels across participants and is reduced to rank $128$ by PCA. We then test syntactic ability on BLiMP \citep{warstadt2020}, a benchmark of English minimal pairs covering eight grammatical phenomena, none of which appear in the fine-tuning text.

We train for six epochs with AdamW at learning rate $5\times10^{-5}$, weight decay $0.01$, batch size $16$, alignment weight $\lambda{=}1$, aligning hidden-state layer $6$ unless stated otherwise. Four conditions differ only in what the model is aligned to; optimization, initialization and data order are identical. \emph{LM-only} switches the alignment off ($\lambda{=}0$). \emph{Brain-aligned} uses the real fMRI targets. \emph{Shuffled} permutes those targets across sentences, keeping their statistics but destroying the match to each sentence. \emph{Random} replaces them with noise of the same dimension, keeping the training pressure but removing the brain entirely. The design attributes a benefit to brain correspondence only if brain-aligned beats \emph{both} controls.

\textbf{Alignment raises CKA to the fMRI target.} End-of-training CKA between the aligned layer and the fMRI target rises from $0.095$ without the alignment loss to $0.337$ with it (Table~\ref{tab:brain} rounds these to two decimals). This is measured on the same $627$ sentences the model was fine-tuned on, so it states how far the objective moved the geometry it was optimizing, and not how well the alignment generalizes to unseen stimuli. Every condition is scored the same way, so the comparison between them is unaffected. A subject-split noise ceiling, the split-half CKA between two independent halves of participants built with the identical rank-128 construction, is $\approx 0.54$, far above the raw-voxel reliability of $0.11$. Read on its own, the objective reached $62$--$78\%$ of the attainable reliable structure across scales, and a paper reporting only this number would be reporting a large effect.

\begin{figure}[t]
\centering
\includegraphics[width=0.86\linewidth]{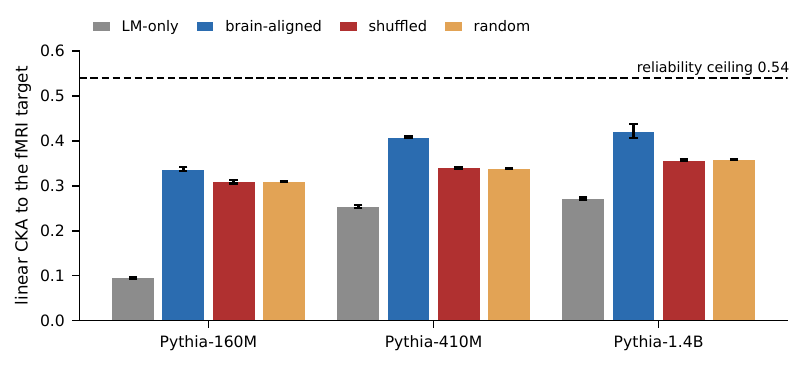}
\caption{What the alignment score reads once the brain is removed from the target. Bars are mean end-of-training CKA to the fMRI target with 95\% confidence intervals over seeds (160M: 20; 410M/1.4B: 8). \emph{Shuffled} permutes targets across sentences and \emph{random} draws rank-matched Gaussian targets, so both retain the second-order structure and regularization pressure of the real target while carrying no correspondence to it. They recover $83$--$92\%$ of the brain-aligned score. The dashed line is the subject-split reliability ceiling; the brain-specific increment is the gap between the blue bar and the two beside it, not the gap to LM-only.}
\label{fig:cka}
\end{figure}

\textbf{Broken targets recover most of that score.} The content ablation establishes how much of that figure is attributable to the brain (Figure~\ref{fig:cka}). Targets permuted across sentences reach $0.31$, and rank-matched Gaussian noise reaches $0.31$ as well: between $83$ and $92\%$ of the brain-aligned score is recovered by a target whose correspondence to the brain has been destroyed. What is attributable to brain content is the increment over those controls, $+0.028$, $+0.068$ and $+0.064$ at 160M, 410M and 1.4B (paired over shared seeds, sign-flip $p<0.01$ at every scale), or $5$--$13\%$ of the ceiling. The increment is reliably nonzero, so brain-specific structure was installed rather than merely attempted; it is also small enough that the score is dominated by something else.

\textbf{Target against target.} Two further facts locate that something else, one about the measure and one about the training, and they should not be added together. Compare the targets to each other first, with no model involved. A shuffled target scores $\mathrm{CKA}=0.204$ against the real one, and a noise target $0.186$. The measure returns a large number when there is no correspondence at all, and the figure follows from the rank. Sweeping the target rank gives $0.051$, $0.102$, $0.204$ and $0.409$ at ranks $32$, $64$, $128$ and $256$, which match $k/n$ for $k$ the rank and $n{=}627$ the number of sentences to within $1\%$ at every point. The pattern is what one would expect of two rank-$k$ subspaces of an $n$-dimensional sample space, which overlap by about $k/n$ whether or not either carries the reference structure. We verified it at four ranks on one target, so we report it as a measured regularity of this construction rather than a proven identity. It may also depend on the target's spectrum, not only on its nominal rank. In practice: report $k/n$ beside any CKA scored against a rank-$k$ target, and score a shuffled target of the same rank before training anything. A rank chosen to keep most of the variance can set a large baseline: at rank $256$ here it would be $0.41$.

\textbf{Model against target.} Between model and target, the movement is almost entirely generic: an unaligned model sits at $0.095$, pulling it toward any target of this construction raises that to $0.31$, and brain content adds the final $0.028$. These two facts do not add up into a single decomposition, and the reason is visible in the numbers: the unaligned model sits at $0.095$, below the $0.204$ that two targets score against each other. A model's geometry is not a target matrix, so none of its score should be charged to that baseline arithmetically, and the baseline is not a lower bound on model scores. We use the ceiling as a scale for model scores while declining to use the baseline, and the asymmetry is deliberate: the two are different kinds of quantity. The ceiling is a \emph{reliability} bound, the share of the target that is reproducible across independent participants, and nothing measured against a target can exceed the part of it that is stable, whatever the measuring representation is made of. The baseline is a \emph{similarity} between two matrices built the same way, and it therefore applies only to comparisons of that kind.

\textbf{What a percentage of the ceiling means.} The baseline cuts against our own framing in one respect. The ceiling is itself a target-to-target comparison of the same construction, so it too sits above the $k/n$ baseline and not above zero. A percentage of the ceiling is therefore a proportion of a quantity that has its own baseline built in, and is not a percentage of attainable signal in an absolute sense. That is a further reason we report raw CKA alongside every such figure, and readers who prefer to avoid the question entirely can use those. The practical consequence: a CKA magnitude reported without an ablated control beside it is close to uninformative about whether the reference structure is the brain. The baseline's rank dependence we have characterized; how the model-side recovery fraction moves with rank would need the models retrained at each rank, which we have not done.

\textbf{No gain on BLiMP.} The downstream result points the same way: brain alignment does not improve BLiMP. At 160M over twenty seeds it reaches $74.5\%$ against $74.1\%$ for LM-only, a $0.48$-point difference that is not significant ($p{=}0.36$). It is not significantly above the shuffled ($74.2\%$, $p{=}0.57$) or random ($73.6\%$, $p{=}0.15$) controls either. The pattern repeats at 410M and 1.4B (Table~\ref{tab:brain}); brain-aligned is never significantly above LM-only ($p{=}0.36,0.29,0.55$), and at 1.4B both controls sit numerically above it. An equivalence test, which asks how large an effect the data can still rule out, bounds the true 160M effect within $[-0.4,+1.4]$ BLiMP points (and within $[-0.4,+1.6]$, $[-1.0,+2.0]$ at the larger scales).

Four explanations for a null are ruled out. The target is not too noisy, which the reliability ceiling establishes. The alignment weight is not mis-tuned: pushing $\lambda$ to $10$ raises CKA to $0.38$ and still buys no detectable gain (Appendix~\ref{app:brain}). The models are not too small, since three scales behave alike. And the training regime is not inert, since fine-tuning moves BLiMP by $3$--$6$ points (Table~\ref{tab:brain}). That last movement is downward, but all four conditions share it, so it cancels in the comparison between them. The null bounds the utility of CKA-to-brain alignment for this operationalization: one fMRI dataset of $627$ sentences, one rank-128 mean-pooled target and one similarity objective. Within it, the objective did most of what such an objective does and bought no detectable gain. It bounds the utility of \emph{brain content specifically} more loosely, because only $5$--$13\%$ of the ceiling was brain-specific to begin with. Both readings point the same way for a practitioner considering the loss, and they differ only for a reader asking about the brain instead of the method.

\begin{table}[t]
\centering
\caption{Brain-alignment study. BLiMP accuracy (\%) for the four conditions at three scales (160M: 20 seeds; 410M/1.4B: 8 seeds), the brain-vs-LM paired $p$, and CKA to the fMRI target for every condition (the content ablation; ceiling $\approx0.54$). Shuffled and random CKA agree to within $0.002$ everywhere and are shown as one column. $\Delta$ is the brain-specific increment, brain minus shuffled, paired over shared seeds. Across-seed standard deviations are $0.8$--$2.0$ BLiMP points and $0.002$--$0.022$ CKA for every cell; at eight seeds a two-sided sign-flip test cannot fall below $p{=}0.0078$, so the $410$M and $1.4$B increments are reported as reliably nonzero and not as precisely estimated. Brain alignment is never significantly above LM-only and never reliably above the controls, while demonstrably reshaping geometry, most of which the geometry-broken targets reproduce.}
\label{tab:brain}
\small
\setlength{\tabcolsep}{4pt}
\begin{tabular}{lccccccccc}
\toprule
& \multicolumn{5}{c}{BLiMP accuracy (\%)} & \multicolumn{4}{c}{CKA to fMRI target} \\
\cmidrule(lr){2-6}\cmidrule(lr){7-10}
Model & LM & Brain & Shuf & Rand & $p$(br.\ vs.\ LM) & LM & Brain & Shuf/Rand & $\Delta$ \\
\midrule
Pythia-160M & 74.1 & 74.5 & 74.2 & 73.6 & 0.36 & 0.10 & 0.34 & 0.31/0.31 & $+0.028$ \\
Pythia-410M & 81.6 & 82.2 & 81.4 & 81.1 & 0.29 & 0.25 & 0.41 & 0.34/0.34 & $+0.068$ \\
Pythia-1.4B & 82.3 & 82.8 & 83.6 & 83.7 & 0.55 & 0.27 & 0.42 & 0.36/0.36 & $+0.064$ \\
\midrule
\multicolumn{10}{l}{\emph{Pretrained, no fine-tuning:} 76.9 (160M), 86.4 (410M), 88.0 (1.4B) $\Rightarrow$ fine-tuning lowers BLiMP $3$--$6$ pts.}\\
\bottomrule
\end{tabular}
\end{table}

\section{Discussion}\label{sec:discussion}
The two studies are methodologically unrelated and use different measures, yet the audit returns the same verdict in both: a large share of the correspondence score is not the correspondence. A permuted fMRI target reproduces most of an alignment score, and $64$ of $272$ transfer estimates in a typological gradient are read off a probe that sits at chance. In neither case is the underlying phenomenon absent. Brain-specific structure was installed, and transfer genuinely decays with distance within the languages where it can be measured. What is absent is the proportionality that makes a score readable at face value, and in the cross-lingual case even the size of the shortfall cannot be pinned down, because the contaminant and the predictor are collinear by construction.

Both failures are easy to miss for the same reason. In each case the contaminating quantity is correlated with the thing of interest rather than independent of it. Second-order structure travels with a real fMRI target; probe reliability travels with typological proximity. A contaminant that co-varies with the signal does not look like noise in a scatter plot, it looks like a stronger effect, and it inflates rather than obscures. That is also why the standard defense, reporting a confidence interval on the score, does not help: the interval is tight around a number that is measuring the wrong thing. The cross-lingual study is a cautionary case in point twice over. Without the random-direction baseline, one could have reported that steering weakens with typological distance, when the pattern is really that unusual languages are easier to disturb by anything. And had we counted pairs instead of languages, that same result would have carried $p=0.0006$, where the correct null gives $0.155$. We make no survey claim about how often this happens in published work; we report that our own analysis was one permutation choice away from it, and that the choice is easy to make without noticing. The steering null is the one place where the intervention was shown to work before its variation was tested, which makes its flatness informative rather than empty.

\textbf{Limitations.} Both studies bound the effect of \emph{specific} operationalizations. The brain-alignment study's target is a sentence-level, mean-pooled, rank-128 fMRI representation; a target constructed to preserve syntactic structure specifically could behave differently, and our noise ceiling establishes reliability, not syntactic relevance. The cross-lingual study uses one model, one agreement phenomenon, one probe family, and a single layer and token position, chosen in advance as the middle of the network and the agreeing verb rather than selected by sweeping. A different layer or position might yield working probes in the four languages that fail here, which would change what the instrument result is about; richer, non-linear interventions might also carry information a linear direction does not. The language-level permutation treats languages as exchangeable, which family structure (Hindi and Urdu, the Romance group) does not fully allow, and ignores that ordered pairs are asymmetric between source and target; a crossed source-by-target model or a family-stratified permutation would be a stronger check. The random steering direction matches the real one in size but not in how it sits relative to the activations; directions drawn from the span of trained probes, or probes trained on shuffled labels, would be stricter controls. The two studies also use different models, chosen for language coverage and for three sizes trained on the same data, so neither result is shown for one model in both settings. The natural next step is a positive control, an alignment target or steering direction with an antecedently expected benefit, which would calibrate the size of effect this design resolves directly, instead of through the equivalence bounds we report.

The four chance-level probes are ours, and we do not identify why they fail. Training-set size is ruled out above. The remaining candidates we cannot separate here: the model's pretraining share for these languages, tokenization of morphologically rich forms, or a genuine difference in how subject-verb agreement is encoded in agglutinative and templatic morphology as against the fusional languages that dominate the sample. But the recommendation does not depend on which is right. Whatever makes a probe fail within a language, a transfer accuracy measured from it is uninterpretable, and a gradient fitted across languages whose probes differ in reliability mixes two quantities regardless of the reason they differ. Indeed the confound is worse if the cause is pretraining share or corpus size, since those correlate with typological distance from English for reasons that have nothing to do with syntax. The check is cheap, it is already computed in any transfer study that fits per-language probes, and reporting it costs a column.

\section{Conclusion}
A permuted or noise target reproduces $83$--$92\%$ of a brain-alignment score, leaving a brain-specific increment of $5$--$13\%$ of the reliability ceiling. A cross-lingual typology gradient is measured, for four of seventeen languages, through probes at or below chance, and because probe reliability declines along the same axis as the predictor, no exclusion this design permits separates the gradient from its own measurement error. Counting seventeen languages rather than $272$ pairs removes the significance of the causal leg. The underlying phenomena are real in both settings; their raw magnitudes are not supported. Both interventions work, and we detect no downstream or distance-graded benefit from what survives. With seventeen languages and eight to twenty seeds, these are failures to detect an effect of the size our designs resolve, not proofs of absence, and they hold for these operationalizations only. Two cases are not a general law, but the checks that produced these numbers are cheap, none required an experiment the original studies had not already run, and each was decisive about what its score meant.
\begin{ack}
The author thanks Dr.\ Sidney Fels (University of British Columbia) for access to computing resources of the Digital Research Alliance of Canada. No specific grant funding was received.
\end{ack}

\bibliography{references}

\begin{thebibliography}{21}
\providecommand{\natexlab}[1]{#1}
\providecommand{\url}[1]{\texttt{#1}}
\expandafter\ifx\csname urlstyle\endcsname\relax
  \providecommand{\doi}[1]{doi: #1}\else
  \providecommand{\doi}{doi: \begingroup \urlstyle{rm}\Url}\fi

\bibitem[Biderman et~al.(2023)Biderman, Schoelkopf, Anthony, Bradley, O'Brien,
  Hallahan, Khan, Purohit, Prashanth, Raff, Skowron, Sutawika, and van~der
  Wal]{biderman2023}
S.~Biderman, H.~Schoelkopf, Q.~Anthony, H.~Bradley, K.~O'Brien, E.~Hallahan,
  M.~A. Khan, S.~Purohit, U.~S. Prashanth, E.~Raff, A.~Skowron, L.~Sutawika,
  and O.~van~der Wal.
\newblock {Pythia}: A suite for analyzing large language models across training
  and scaling.
\newblock In \emph{ICML}, 2023.

\bibitem[Dijujin et~al.(2025)Dijujin, Rohani, Samiei, and
  Baghshah]{Dijujin2025}
N.~H. Dijujin, S.~R.~R. Rohani, M.~M. Samiei, and M.~S. Baghshah.
\newblock Inductive biases for zero-shot systematic generalization in
  language-informed reinforcement learning.
\newblock \emph{Machine Learning}, 2025.
\newblock \doi{10.1007/s10994-025-06764-7}.

\bibitem[Feinman and Lake(2018)]{Feinman2018}
R.~Feinman and B.~Lake.
\newblock Learning inductive biases with simple neural networks.
\newblock \emph{Annual Meeting of the Cognitive Science Society}, 2018.

\bibitem[Hewitt and Liang(2019)]{hewitt2019}
J.~Hewitt and P.~Liang.
\newblock Designing and interpreting probes with control tasks.
\newblock In \emph{Proceedings of the 2019 Conference on Empirical Methods in
  Natural Language Processing}, 2019.

\bibitem[Jumelet et~al.(2025)Jumelet, Weissweiler, and Bisazza]{jumelet2025}
J.~Jumelet, L.~Weissweiler, and A.~Bisazza.
\newblock {MultiBLiMP} 1.0: A massively multilingual benchmark of linguistic
  minimal pairs, 2025.

\bibitem[Kornblith et~al.(2019)Kornblith, Norouzi, Lee, and
  Hinton]{kornblith2019}
S.~Kornblith, M.~Norouzi, H.~Lee, and G.~Hinton.
\newblock Similarity of neural network representations revisited.
\newblock In \emph{Proceedings of the 36th International Conference on Machine
  Learning}, 2019.

\bibitem[Lin et~al.(2021)Lin, Mihaylov, Artetxe, Wang, Chen, Simig, Ott, Goyal,
  Bhosale, Du, Pasunuru, Shleifer, Koura, Chaudhary, O'Horo, Wang, Zettlemoyer,
  Kozareva, Diab, Stoyanov, and Li]{lin2021xglm}
X.~V. Lin, T.~Mihaylov, M.~Artetxe, T.~Wang, S.~Chen, D.~Simig, M.~Ott,
  N.~Goyal, S.~Bhosale, J.~Du, R.~Pasunuru, S.~Shleifer, P.~S. Koura,
  V.~Chaudhary, B.~O'Horo, J.~Wang, L.~Zettlemoyer, Z.~Kozareva, M.~Diab,
  V.~Stoyanov, and X.~Li.
\newblock Few-shot learning with multilingual language models, 2021.

\bibitem[Lin et~al.(2019)Lin, Chen, Lee, Li, Zhang, Xia, Rijhwani, He, Zhang,
  Ma, Anastasopoulos, Littell, and Neubig]{lin2019}
Y.-H. Lin, C.-Y. Chen, J.~Lee, Z.~Li, Y.~Zhang, M.~Xia, S.~Rijhwani, J.~He,
  Z.~Zhang, X.~Ma, A.~Anastasopoulos, P.~Littell, and G.~Neubig.
\newblock Choosing transfer languages for cross-lingual learning.
\newblock In \emph{ACL}, 2019.

\bibitem[Littell et~al.(2017)Littell, Mortensen, Lin, Kairis, Turner, and
  Levin]{littell2017}
P.~Littell, D.~R. Mortensen, K.~Lin, K.~Kairis, C.~Turner, and L.~Levin.
\newblock {URIEL} and lang2vec: Representing languages as typological,
  geographical, and phylogenetic vectors.
\newblock In \emph{EACL}, 2017.
\newblock \doi{10.18653/v1/E17-2002}.

\bibitem[Mantel(1967)]{mantel1967}
N.~Mantel.
\newblock The detection of disease clustering and a generalized regression
  approach.
\newblock \emph{Cancer Research}, 27\penalty0 (2):\penalty0 209--220, 1967.

\bibitem[Mueller et~al.(2020)Mueller, Nicolai, Petrou-Zeniou, Talmina, and
  Linzen]{mueller2020}
A.~Mueller, G.~Nicolai, P.~Petrou-Zeniou, N.~Talmina, and T.~Linzen.
\newblock Cross-linguistic syntactic evaluation of word prediction models.
\newblock In \emph{ACL}, 2020.

\bibitem[Papadimitriou and Jurafsky(2023)]{Papadimitriou2023}
I.~Papadimitriou and D.~Jurafsky.
\newblock Injecting structural hints: Using language models to study inductive
  biases in language learning, 2023.

\bibitem[Pereira et~al.(2018)Pereira, Lou, Pritchett, Ritter, Gershman,
  Kanwisher, Botvinick, and Fedorenko]{pereira2018}
F.~Pereira, B.~Lou, B.~Pritchett, S.~Ritter, S.~J. Gershman, N.~Kanwisher,
  M.~Botvinick, and E.~Fedorenko.
\newblock Toward a universal decoder of linguistic meaning from brain
  activation.
\newblock \emph{Nature Communications}, 2018.
\newblock \doi{10.1038/s41467-018-03068-4}.

\bibitem[Pezzulo et~al.(2026)Pezzulo, Nuzzi, D'Alessandro, Proietti, Bottini,
  and Cisek]{Pezzulo2026}
G.~Pezzulo, D.~Nuzzi, M.~D'Alessandro, R.~Proietti, R.~Bottini, and P.~Cisek.
\newblock Grounded world models in biological organisms and future embodied
  {AI}, 2026.

\bibitem[Rahbar(2026{\natexlab{a}})]{Rahbar2026curse}
S.~Rahbar.
\newblock {The Curse of Multilinguality in Lexical Normalization}.
\newblock \emph{arXiv preprint arXiv:2609.00329}, 2026{\natexlab{a}}.
\newblock \doi{10.48550/arXiv.2609.00329}.

\bibitem[Rahbar(2026{\natexlab{b}})]{Rahbar2026ignition}
S.~Rahbar.
\newblock {The Ignition Index: Measuring Global Workspace Dynamics in Language
  Models}.
\newblock \emph{arXiv preprint arXiv:2608.05160}, 2026{\natexlab{b}}.
\newblock \doi{10.48550/arXiv.2608.05160}.

\bibitem[Rahbar(2026{\natexlab{c}})]{Rahbar2026site}
S.~Rahbar.
\newblock {Site Is Decodable Before Pretraining: Negative Controls for Probing
  Frozen Brain-MRI Foundation Models}.
\newblock \emph{arXiv preprint arXiv:2608.10295}, 2026{\natexlab{c}}.
\newblock \doi{10.48550/arXiv.2608.10295}.

\bibitem[Turner et~al.(2023)Turner, Thiergart, Leech, Udell, Vazquez, Mini, and
  MacDiarmid]{turner2023}
A.~M. Turner, L.~Thiergart, G.~Leech, D.~Udell, J.~J. Vazquez, U.~Mini, and
  M.~MacDiarmid.
\newblock Steering language models with activation engineering, 2023.

\bibitem[Warstadt et~al.(2020)Warstadt, Parrish, Liu, Mohananey, Peng, Wang,
  and Bowman]{warstadt2020}
A.~Warstadt, A.~Parrish, H.~Liu, A.~Mohananey, W.~Peng, S.-F. Wang, and S.~R.
  Bowman.
\newblock {BLiMP}: The benchmark of linguistic minimal pairs for {English}.
\newblock \emph{Transactions of the Association for Computational Linguistics},
  2020.

\bibitem[Zhang et~al.(2026)Zhang, Chai, Wu, Siok, and Wang]{Zhang2026}
F.~Zhang, B.~Chai, Y.~Wu, W.~Siok, and N.~Wang.
\newblock Linguistics and human brain: a perspective of computational
  neuroscience.
\newblock \emph{Cognitive Neurodynamics}, 2026.
\newblock \doi{10.1007/s11571-026-10496-2}.

\bibitem[Zhao et~al.(2025)Zhao, Wu, Yang, Zou, and Hong]{Zhao2025}
H.~Zhao, H.~Wu, D.~Yang, A.~Zou, and J.~Hong.
\newblock {BriLLM}: Brain-inspired large language model, 2025.

\end{thebibliography}

\bibliographystyle{abbrvnat}

\appendix
\section{Cross-lingual study: details}
\label{app:xling}
\textbf{Languages (17):} English, German, French, Spanish, Italian, Portuguese, Russian, Bulgarian, Greek, Finnish, Estonian, Turkish, Catalan, Hindi, Urdu, Basque, Arabic (MultiBLiMP $\cap$ model support). Within-language probe accuracy averages $0.729$ (range $0.375$ Hindi to $0.901$ Finnish); two of the $272$ transfer pairs exceed the highest within-language accuracy.

\textbf{Training size is held fixed by design.} MultiBLiMP supplies between $273$ (Basque) and $3{,}832$ (Russian) usable minimal pairs per language, but we cap every language at $300$ and train the probe on $60\%$ of them. Sixteen of the seventeen therefore fit on exactly $180$ pairs and Basque on $163$. Within-language accuracy correlates with the number actually used at $r=-0.04$; the apparent $r=+0.42$ against the number \emph{available} is an artifact of availability being irrelevant beyond the cap. Basque, with the least training data of any language, reaches $0.755$, above all four chance-level languages.

\textbf{Distance} is cosine distance over lang2vec \texttt{syntax\_knn} URIEL features (e.g.\ eng--deu $0.10$, eng--tur $0.49$).

\textbf{Significance} is by permutation over \emph{language labels} ($10{,}000$ draws), the Mantel construction, rather than over the $272$ pairs: each language appears in $32$ pairs and URIEL distances are phylogenetically structured, so pairs are not exchangeable and a pair-level null overstates significance. The confound-corrected leg uses a per-target random-direction baseline averaged over twenty random directions; the decisive control is a partial correlation of the corrected steering benefit with distance, controlling for probe-transfer accuracy.

\textbf{Summary,} with pair-level $p$ in brackets for comparison: transfer$\sim$distance $r=-0.66,\,p{=}10^{-4}$ [$2{\times}10^{-4}$], and $r=-0.47,\,p{=}0.025$ restricted to the thirteen above-chance-probe languages; raw steering$\sim$distance $r=+0.08,\,p{=}0.586$ [$0.18$]; random-direction null $r=+0.24,\,p{=}0.081$ [$2{\times}10^{-4}$]; corrected steering$\sim$distance $r=-0.22,\,p{=}0.155$ [$0.0006$], and $r=-0.09,\,p{=}0.704$ on the above-chance-probe subset; partial (corrected $\mid$ transfer) $r=+0.11,\,p{=}0.455$ [$0.072$]. Only the transfer gradient survives the language-level null, and the corrected steering leg is the clearest case of the gap: strongly significant on pairs, null on languages. The within-language manipulation check gives a mean corrected margin shift of $+6.21$, positive in 16 of 17 languages, sign-flip $p \le 5\times10^{-5}$ at $20{,}000$ draws.

\textbf{Probe-validity robustness.} The exclusion in Section~\ref{sec:xling} uses a threshold of $0.60$ on within-language accuracy, but nothing depends on that value: thresholds of $0.50$, $0.525$, $0.55$ and $0.60$ all remove the same four languages ($r=-0.468$, $156$ pairs), and $0.65$ to $0.75$ additionally remove German ($r=-0.455$, $132$ pairs). Only at $0.80$ does the estimate move materially, and by then eight languages and $56$ pairs remain, which is over-trimming, not a threshold effect. To test whether the attenuation is specific to these languages and not a consequence of dropping any four, we enumerated all $\binom{17}{4}=2{,}380$ four-language exclusions: their $r$ values have mean $-0.654$ and standard deviation $0.048$, and only two of the $2{,}380$ attenuate the correlation more than removing the chance-level set does. That figure looks like strong specificity evidence and is not, for the reason given immediately below; we report it because it is the test a reader would expect and because its failure is the point. A leave-one-language-out jackknife shows no single language carries the effect: the largest weakening comes from removing Turkish, at $+0.029$, and removing Finnish or English \emph{strengthens} it. Within the four excluded languages alone the gradient is also present ($r=-0.536$, $12$ pairs), so the relationship is not confined to the languages we keep.

\textbf{Range restriction, and why we do not quantify the instrument's share.} The exclusion enumeration above is not the test it first appears to be. Mean typological distance to the rest of the sample orders the languages Turkish $0.397$, Basque $0.310$, Urdu $0.307$, Hindi $0.302$, Arabic $0.302$, then a gap to English $0.247$ and below, so the four chance-level languages occupy ranks one, three, four and five. Dropping them truncates the predictor, and most of the $2{,}380$ alternative exclusions remove central languages and restrict no range at all; the enumeration therefore contrasts a range-restricting exclusion against a null dominated by non-restricting ones. Conditioning the null on the removed set's summed mean distance leaves only three other sets at least as extreme as the invalid one ($1.307$): two attenuate the correlation more ($r=-0.458$ and $-0.331$) and the third about as much ($-0.484$, against $-0.468$). The four most distant \emph{valid} languages sum to $1.047$, so no distance-matched exclusion exists in this sample. We therefore report the $r=-0.47$ figure as what the gradient becomes when the unmeasurable languages are removed, and explicitly not as an estimate of how much of the gradient the instrument contributed. Separating the two needs a sample in which probe reliability and typological distance are not collinear, which is a sampling problem rather than an analysis one.

\textbf{Precision of the within-language estimates.} Each within-language accuracy is the mean over five train/test resamples, but the resamples draw from the same capped $300$ pairs, so they do not give five independent estimates; the binding uncertainty is the $\approx 120$-item test set, a binomial standard error of $4.6$ points at chance. The four excluded languages have $95\%$ intervals of $[0.286,0.464]$ (Hindi), $[0.344,0.523]$ (Urdu), $[0.403,0.582]$ (Turkish) and $[0.407,0.586]$ (Arabic); all four exclude the $0.60$ threshold, so the exclusion decision is not marginal even though the point estimates are imprecise.

One estimate is anomalous in a way imprecision does not explain: Hindi at $0.375$ is $2.7$ standard errors \emph{below} chance ($p=0.006$), where Urdu, Turkish and Arabic are statistically at chance ($p=0.14$, $0.87$, $0.94$). A probe reliably below chance on held-out data is more consistent with inverted labels or a mislocated critical token, for a verb-final language, than with difficulty. We flag it rather than resolve it, and we check what it would cost if it were a fault. Dropping Hindi entirely leaves the reliability gradient at $r=-0.745$ against $-0.743$ with it, the full-cohort transfer gradient at $-0.664$ against $-0.657$, and the thirteen-language fit unchanged at $-0.468$ since Hindi is excluded from it in any case. Every relationship the paper reports is therefore robust to removing the one language whose estimate we distrust.

\section{Brain-alignment study: details}
\label{app:brain}
\textbf{Alignment-weight sweep (160M, 20 seeds).} $\lambda{=}0.1$: BLiMP $73.6$, CKA $0.18$; $\lambda{=}1$: $74.5$, $0.34$; $\lambda{=}10$: $73.8$, $0.38$; LM-only $74.1$, CKA $0.10$. Stronger alignment raises CKA monotonically with no BLiMP gain.

\textbf{Layer ablation (160M).} Early/middle/late aligned layer: $74.9/74.5/73.5$, all within a point of the $74.1$ LM-only score.

\textbf{Statistics.} Paired $t$-tests over matched seeds; $95\%$ CIs from the $t$-distribution; TOST $90\%$ CIs $[-0.4,+1.4]$ (160M), $[-0.4,+1.6]$ (410M), $[-1.0,+2.0]$ (1.4B). All code, including the script that builds the rank-128 targets from the public Pereira et al.\ data, and the raw results for both studies are available at \url{https://github.com/saman-rahbar/score-is-not-the-structure}.

\textbf{CKA estimator.} The soft linear-CKA estimator column-centers both matrices first, $\tilde{H}=H-\bar{H}$ and $\tilde{B}=B-\bar{B}$ with the mean taken over the batch dimension, then computes $\operatorname{CKA}(H,B)=\lVert \tilde{H}^\top \tilde{B}\rVert_F^2 / (\lVert \tilde{H}^\top \tilde{H}\rVert_F \lVert \tilde{B}^\top \tilde{B}\rVert_F)$, with a small constant added to the denominator for numerical stability. The centering matters for reading this paper: an uncentered variant is inflated by shared means, which is one mechanism that could by itself produce a high score for the ablated targets, so the recovery figures we report are not that artifact. All models are held in single precision.

\textbf{Power at scale.} The 410M and 1.4B comparisons use eight seeds and are correspondingly lower-powered than the twenty-seed 160M anchor; their TOST intervals above are wider, and at 1.4B the controls sit numerically above the brain-aligned condition. We therefore treat the larger scales as consistent bounded replications (no positive effect detected) and not as independent high-powered equivalence tests, and anchor the equivalence claim at 160M.

\textbf{The target-to-target baseline.} Scoring targets against each other instead of against a model isolates how much CKA is available without any correspondence. Using the same rank-128 construction, the same permutation, and the same random draw the experiment uses, and averaging over twenty draws: $\mathrm{CKA}(\text{true},\text{true})=1.000$ by construction, $\mathrm{CKA}(\text{true},\text{shuffled})=0.204$ (sd $0.001$), $\mathrm{CKA}(\text{true},\text{random})=0.186$ (sd $0.002$). Two consequences, both stated for the comparison they belong to. Among matrices of this construction, a value near $0.2$ carries no correspondence information at all, so comparisons of that kind run from roughly $0.2$ to the $0.54$ ceiling and not from zero. The model-to-target comparison is a different one and does not inherit that baseline: the unaligned model sits at $0.095$, below it, because a layer's geometry is not a rank-128 target matrix. What the baseline does establish is that this measure is capable of returning a large number in the complete absence of correspondence, which is the property the content ablation exposes in the trained conditions. Separately, since a model trained on an ablated target reaches $0.31$, above the $0.20$ its own target sits at, those conditions are not merely inheriting their target's similarity to the true one; the alignment objective moves any model toward the true target's geometry whether or not the target carries it. These figures need only the target matrix and no model of any kind, and are reproduced by the released code.

Sweeping the rank gives $\mathrm{CKA}(\text{true},\text{shuffled})$ of $0.0505$, $0.1019$, $0.2040$ and $0.4088$ at ranks $32$, $64$, $128$ and $256$, against $k/n$ of $0.0510$, $0.1021$, $0.2041$ and $0.4083$: a ratio of $0.99$ to $1.00$ throughout. The rank-matched-noise column tracks $k/n$ less exactly ($0.99$ down to $0.84$ as rank grows), as expected, since a Gaussian draw is not a permutation of the target and so does not preserve its column space. The consequence for reading any alignment score is that the rank of the target sets the baseline, and the choice is usually made for variance reasons with no reference to what it does to the baseline.

\textbf{fMRI target construction.} We use the two Pereira et al.\ (2018) sentence-reading experiments via the Brain-Score language package, which distributes responses already restricted to language-network voxels: the 243-sentence experiment (six participants, 8{,}031 voxels) and the 384-sentence experiment (nine participants, 12{,}155 voxels), $627$ sentences total. Participants overlap between the two experiments, so a voxel can be identified across them by its full index; we keep the voxels present in both and stack the experiments into a single $627$-sentence matrix, rather than analyzing them separately or averaging across participants. Each retained voxel is then z-scored across sentences, and the top-128 principal components of the result are the target. The noise ceiling splits the participants of each experiment into two random halves (50 random 3-vs-3 / 4-vs-5 splits), builds each half's rank-128 target, and computes the linear CKA between the two halves' sentence Gram structure (which is well-defined across the halves' different voxel sets); the raw-voxel figure ($0.11$) is the same split-half CKA computed on the un-reduced voxel responses. The $0.54$ headline is the sentence-count-weighted mean across the two experiments.

\end{document}